# Comparative Analysis of Shear Strength Prediction Models for Reinforced Concrete Slab-Column Connections

**Sarmed Wahab[1], Nasim Shakouri Mahmoudabadi[2], Sarmad Waqas[3], Nouman Herl[4], Muhammad Iqbal[5], Khurshid Alam[6], Afaq Ahmad[7]**

1. Civil Engineering Department, University of Engineering and Technology, Taxila, Rawalpindi, Pakistan, 21-ms-str-3@students.uettaxila.edu.pk
2. Department of Civil Engineering, The University of Memphis, TN 38152, USA; nshkrmhm@memphis.edu;
3. Project Manger. Osmani & Compnay Pvt. Ltd. Pakistan  sarmad.waqas@gmail.com
4. Federal Board of Intermediate and Secondary Education, Pakistan noumanherl@hotmail.com
5. Department of Mechanical Engineering, CECOS University of IT & Emerging Sciences, Hayatabad Peshawar 25000, Pakistan; muhammadiqbal@cecos.edu.pk
6. Department of Mechanical and Industrial Engineering, College of Engineering, Sultan Qaboos University, Sultanate of Oman, Oman, kalam@squ.edu.om,
7. Department of Civil Engineering, The University of Memphis, TN 38152, USA; aahmad4@memphis.edu

**\*** Corresponding author: aahmad4@memphis.edu

## Abstract

This research aims at comparative analysis of shear strength prediction at slab-column connection, unifying machine learning, design codes and Finite Element Analysis. Current design codes (CDCs) of ACI 318-19 (ACI), Eurocode 2 (EC2), Compressive Force Path (CFP) method, Feed Forward Neural Network (FNN) based Artificial Neural Network (ANN), PSO-based FNN (PSOFNN), and BAT algorithm-based BATFNN are used. The study is complemented with FEA of slab for validating the experimental results and machine learning predictions. A dataset of 145 simply supported Square Concrete Slab (SCS) samples is compiled from past research work. The parameters involved in the dataset are slab depth ($d_s$), column dimension ($c_s$), slab's shear span ratio ($a_v/d$), yield strength of longitudinal steel ($f_{yls}$), longitudinal reinforcement ratio ($p_{ls}$), ultimate load carrying capacity ($V_{us}$), and concrete compressive strength ($f_{cs}$). In the case of hybrid models of PSOFNN and BATFNN, mean square error is used as an objective function to obtain the optimized values of the weights, that are used by Feed Forward Neural Network to perform predictions on the slab data. Seven different models of PSOFNN, BATFNN, and FNN are trained on this data and the results exhibited that PSOFNN is the best model overall. PSOFNN has the best results for SCS=1 with highest value of R as 99.37% and lowest of MSE, and MAE values of 0.0275%, and 1.214% respectively which are better than the best FNN model for SCS=4 having the values of R, MSE, and MAE as 97.464%, 0.0492%, and 1.43%, respectively. PSOFNN results were even better for the dataset with fewer input parameters as in SCS=5. For the validation of these models, the well know FEA tools i.e.; ABAQUS is used. ABAQUS results provided accurate examination of non-linear behavior of concrete slabs and verified the experimental values and predictions of PSOFNN.

# 1 Introduction

Flat slab is preferred in building reinforced concrete structures because of their economical and efficient construction. It is directly placed over the columns without any beams, which causes direct transfer of the load from the slab to columns. The absence of beams is advantageous in several ways including a reduction in the building's overall height, easy accommodation of vertical shafts, layout flexibility, easy placement of reinforcement, and faster construction. But any beamless slab system including a flat plate has poor resistance against lateral loads such as seismic or wind loads, therefore flat slabs are not preferred in high-rise structures and high seismic areas. Its structural efficiency is hindered by its inadequate performance. Even moderate-intensity earthquake produces excessive damaging deformations. The design and detailing of a flat slab system should consider the seismic hazard level of the region, the expected ground motion, and the characteristics of the building, such as its height, weight, and stiffness. Proper reinforcement detailing, such as using seismic reinforcement and adequate anchorage of the columns to the foundation, is essential for ensuring the system's safety. But if not properly designed the flat slabs can undergo punching shear failure due to large column loads, insufficient concrete strength, inadequate slab thickness, insufficient shear reinforcement, small column heads and poor construction quality.

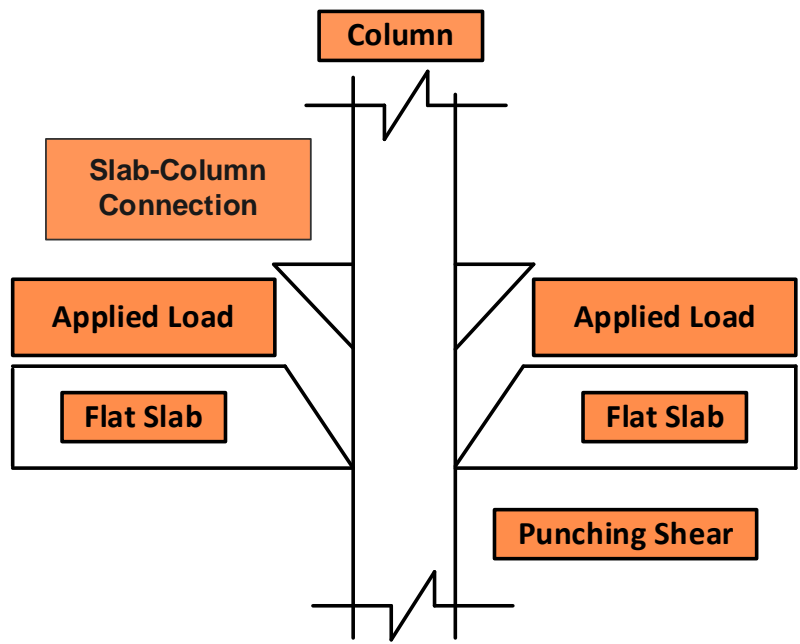

FIGURE 1: Punching shear failure at slab-column connection

These problems develop high shear stress at the connection of the slab and column, as shown in FIGURE 1. However, flat slabs have limited shear strength at the slab-column connection due to absence of beams or drop panels which are present in other types of slab systems, and thus they are vulnerable to punching shear failure. The minimum thickness of a flat slab is often enough for vertical loads but falls short of the high lateral loads developed during seismic activity (Hossen and Anam 2010). The failure of slab in punching shear occurs due to various reasons like unreliable or poor construction, inferior material usage, and excessive loads than the shear capacity. The drop panels play an important role in reducing the punching shear failure but there exist optimum dimensions for them (Yankelevsky, Karinski et al. 2021). A larger contact area of the drop panel does not always mean increased shear resistance. Therefore, the construction of the slab-column joint is of great significance to avoid punching failure. The flat-slab flooring system is widely used in some highly vulnerable seismic regions of the world like Middle Eastern and Mediterranean countries (Erberik and Elnashai 2003).

To meet the above requirements, a unique alternative design approach called Compressive Force Path (CFP) (Kotsovos 2014) is adopted for this research. The Compressive Force Path (CFP) method is a structural design approach that can be used to improve the punching shear strength of flat slabs. The CFP method involves designing the structural elements in such a way that the compressive forces are transmitted efficiently from the slab to the columns, reducing the tensile stresses at the slab-column interface and improving the punching shear strength of the slab. The CFP method involves several design considerations, including the geometry and dimensions of the column, the size and spacing of the reinforcement, and the arrangement of the reinforcement around the column. The method aims to create a force path for the compressive forces that avoids high tensile stresses at the slab-column interface. One of the key design features of the CFP method is the use of a column head that is wider than the column. This wider column head helps to distribute the compressive forces from the slab more efficiently to the column, reducing the tensile stresses at the slab-column interface. The reinforcement design in the CFP method is also critical for improving the punching shear strength of the slab. The reinforcement should be arranged around the column in a circular or square pattern, with an appropriate spacing and size. The reinforcement should be placed close to the surface of the slab to maximize its effectiveness in reducing the tensile stresses. Another important consideration in the CFP method is the use of shear studs or shear connectors to improve the bond between the reinforcement and the concrete. This helps to transfer the compressive forces more efficiently from the slab to the reinforcement, improving the overall strength of the slab. The CFP method is a popular approach to improving the punching shear strength of flat slabs because it is relatively simple to apply and can be used with a variety of structural systems. The method is particularly useful for buildings with large flat slabs and heavy column loads, where punching shear failure is a significant design consideration.

Current Design Codes (CDCs) of ACI-318 (ACI) and Eurocode 2 (EC2) being practiced were developed for limit state design of structures that are focused on providing economical and safer structural designs. This is because the design method is based on the theoretical framework. It was found that these design methods work better for steel but under ultimate load conditions, they are not suited for concrete structures (Kotsovos 2014). CDCs have provided unreliable results for a safer and more economical structural philosophy of limit state design. The experiments performed on structural members like columns, beams, beam-column joints, and walls have shown great variation from the prediction of performance by the CDCs (Kotsovos 2014). In certain cases, the predictions of CDCs overestimate or underestimate the structural performance even when the reinforcement needs, and economic constraints are fulfilled. The CFP method involves two major improvements, the first one involves the determination of the areas in the structural members from where the applied load is transferred from point of its application to supports, and the second includes improving the strength of that area for enhancing structural ductility and load carrying capacity (Kotsovos 2014). The CFP satisfies all code performance requirements at both structural and material levels.

Artificial Neural Networks (ANN) are the advanced computational model of interconnected layers that are at minimum two in the case of single-layer perceptron (Rosenblatt 1958) and three for multilayer perceptron (Werbos 1974) namely the input, hidden, and output layer. These layers contain the neurons that relate to the neurons in the adjacent layer. They require less computational resources and less time for analysis than normal for the finite element (FE) method. ANN is being used in various fields of study in research including design predictions of RC members (Ahmad, Kotsovou et al. 2018), predicting elastic behavior of normal and high strength concrete (Demir 2008), structural behavior of slabs (Hegazy, Tully et al. 1998), ultimate strength of beams (Perera, Barchín et al. 2010), rutting performance of asphalt mixtures containing steel slag aggregates (Shafabakhsh, Ani et al. 2015) and prediction of behavior of shear connectors in concrete (Shariati, Mafipour et al. 2019). ANN is one of the best soft computing methods being used in the research (Waris et al., 2020, Shariati, Mafipour et al. 2019, Shariati, Mafipour et al. 2020, Ghumman et al., 2021), like in structural engineering, they are mostly used to predict the load carrying capacity of structural members like the strength prediction of RC beams (Perera, Tarazona et al. 2014, Shariati, Mafipour et al. 2020, Ahmad et al., 2020), columns (Karimipour, Mohebbi Najm Abad et al. 2021, Haido 2022, Chen, Fakharian et al. 2023). They have been used to predict the load carrying capacity of slabs (Ahmad, Arshid et al. 2021, Faridmehr, Nehdi et al. 2022). They can learn almost any pattern in the data (Anderson and McNeill 1992, Basheer and Hajmeer 2000) and keep a record of it in their memory. The neural networks can be combined with Metaheuristic algorithms to develop hybrid ANN models. Metaheuristic algorithms are nature or physics-based algorithms, developed to perform optimization of an objective function. ANN uses a gradient-based approach for optimization, this method converges faster and provides solutions with high accuracy. However, this type of optimization is not preferred in real-world optimization problems requiring proper consideration of prohibited zones, and boundary conditions for variables. Metaheuristic algorithms are not restricted to these constraints, they consider these conditions while performing the optimization. These methods are effective for exploration (global search) and exploitation (local search) purposes as they do not make assumptions about the original optimization problem (Kaveh 2021). These algorithms are used for the optimization of the research problem by either minimizing or maximizing an objective function (Kennedy and Eberhart 1995, Poli, Kennedy et al. 2007) defined by the user.

The performance of ANN can be optimized using these algorithms. There are numerous ways in which ANN can be optimized including optimizing the architecture of the neural network, weight optimization, activation nodes, and parameters involved in the network (Ojha, Abraham et al. 2017). Several metaheuristic algorithms are developed for optimization purposes, namely Particle Swarm Optimization (PSO), Bat algorithm (BA), Ant Colony, Grey Wolf Optimizer (GWO), and Artificial Bee Colony (ABC). These algorithms include different parameters and decision conditions that affect the complexity and the performance of the algorithm. The major advantages of these algorithms over ANN are their ability to provide the optimum value of weights of the network after performing optimization, and their ability to elude being trapped in local minima and multi-variability (Yang 2011). ANN is based on gradient descent-based algorithms capable of performing the local search. Metaheuristic algorithms combine exploration and exploitation strategies to find a global optimum solution (Osman and Laporte 1996). The conventional algorithms involved in the ANN are fast and perform better in local search, but they lack global search ability, however, the metaheuristic algorithms are effective in global search. Therefore, a combination of the two offers better optimization than an individual of them.

Despite having various methodologies to determine the punching shear strength of flat slabs, these approaches only work under specific framework. Such difficulties in the empirical methods can be avoided by using machine learning algorithms. These hybrid machine learning models have demonstrated their effectiveness in determining the performance of various structural members. A hybrid model of adaptive neuro fuzzy inference systems combined with genetic algorithm, and particle swarm optimization has been used to predict the shear strength of RC beams (Li, Yan et al. 2023). The hybrid model predicted shear strength with greater accuracy compared to standalone models. In another study, an informational Bat ANN was applied to predict the punching shear strength of RC flat slabs without shear reinforcement (Faridmehr, Nehdi et al. 2022), this research investigated 30 distinct topologies of the model to identify the best possible prediction model with minimized errors and highest $R^2$ values. Nolan Concha et al. implemented a hybrid neural network of particle swarm optimization to predict the shear strength of steel fiber reinforced concrete deep beams (Concha, Aratan et al. 2023). The hybrid model predicted the strength of steel fiber reinforced concrete deep beam with a correlation coefficient of 0.997. The high accuracy of these hybrid prediction models provides a suitable tool to predict the structural performance. Sandeep, Tiprak et al. (Sandeep, Tiprak et al. 2023) used machine learning to predict the shear strength of reinforced concrete beams. Researchers used atom search optimization (ASO) algorithm combined with neural network to predict the shear strength of beams. These results were then compared with the prediction results of various hybrid and standalone models including ANN, genetic algorithm, particle swarm optimized neural network, and support vector machines.

The present study is aimed at the prediction of the response of reinforced concrete structural members by comparing the performance of CDCs, CFP, ANN, and hybrid ANN models of PSOFNN and BATFNN. This study aims at predicting the load-carrying capacity of a flat slab by using three different Machine Learning models, one of which is Feed Forward Neural Network (FNN) and the other two hybrid ANN models of FNN based Particle Swarm Optimization (PSOFNN) and Bat algorithm (BATFNN). These models require data for training and testing purposes; therefore 145 data samples are collected from previous research. The data consists of experimental values of SCS under concentric loading. The parameters involved in the dataset are slab depth ($d_s$), column dimension ($c_s$), shear span ratio ($a_v/d$), longitudinal steel yield strength ($f_{yls}$), percentage of longitudinal steel ratio ($p_{ls}$), ultimate load carrying capacity ($V_{us}$), and concrete compressive strength ($f_{cs}$). This data is converted to 7 different datasets using the parameters from the original database. Both hybrid ANN models of PSOFNN and BATFNN were run on the slab data using Mean Square

Error (MSE) objective functions for the optimization of the hybrid models. The optimization function provides the best results for the Mean Square Error, Mean Absolute error (MAE), R-value, and prediction value. These values are then compared with calculated values from CDCs and CFP that revealed that design codes result in smaller values than the original values (Ahmad and Raza 2020, Kotsovou, Ahmad et al. 2020, Raza and Ahmad 2020). PSOFNN performed better than all other prediction models and design codes. Finite Element Modeling (FEM) is performed for validating the result outcomes of the prediction models of ANN and Hybrid ANN. The literature revealed that FEM was used by many researchers (Hamid, Mahdi et al. 2012, Hany, Hantouche et al. 2016, Ibrahim, Fahmy et al. 2016, Alfarah, López-Almansa et al. 2017, Elchalakani, Karrech et al. 2018) and load carrying capacity of materials has also been determined using FEM (Shariati, Ramli Sulong et al. 2010, Hamid, Mahdi et al. 2012). The result of the PSOFNN and ANN resulted in good correspondence with the Finite Element Analysis (FEA) results.

## 2 Current Design Codes

Truss Analogy models of mechanics are used for providing a theory for the transfer of load at the ultimate limit response (ULR) of SCS (Ahmad, Kotsovou et al. 2018). In the truss analogy method, the force is assumed to be transmitted in a triangular fashion. ACI and EC2 equations are empirical by nature (Code 2005, ACI 2014) leading to data fitting, and can cause the failure of structures like collapse (Wood 1997). However, the analysis formula used in the CDCs is of different nature that results in a significant difference between the values calculated using CDCs and the experimental values. However, due to the ease of construction, engineering professionals use CDC for testing the connection of flat slabs and columns at ULR. A few parameters are used by CDC for determining punching failure of slabs that are slab depth ($d_s$), column dimension ($c_s$), shear span ratio ($a_v/d$), longitudinal reinforcement yield strength ($f_{yls}$), longitudinal reinforcement ratio ($p_{ls}$), ultimate load carrying capacity ($V_{us}$), and compressive strength of concrete ($f_{cs}$). The SCS response is predicted using the equation (1) for ACI based boundary conditions as shown in FIGURE 2.

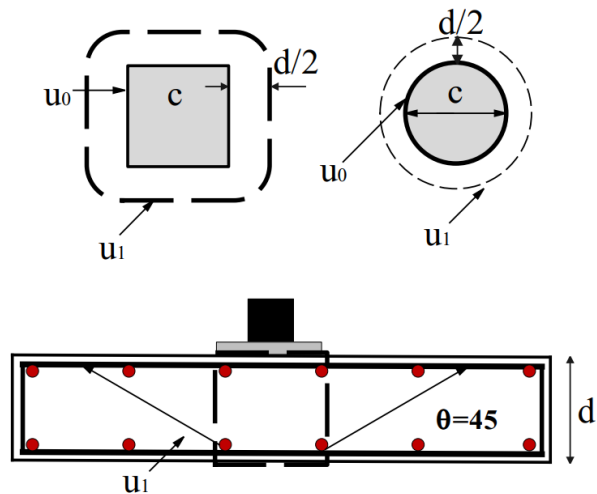

FIGURE 2: ACI-based boundary for SCS

$$V_{rc} = \frac{1}{3} * \lambda_s * u_{2s} * ds * \sqrt{f'_c} \qquad (1)$$

and

$$\lambda_s = \sqrt{\frac{2}{1 + 0.004d}} \leq 1 \qquad (2)$$

$$u_{2s} = 4(c + d) \qquad (3)$$

The following equation is used in EC2 for the prediction of slab response as in FIGURE 3.

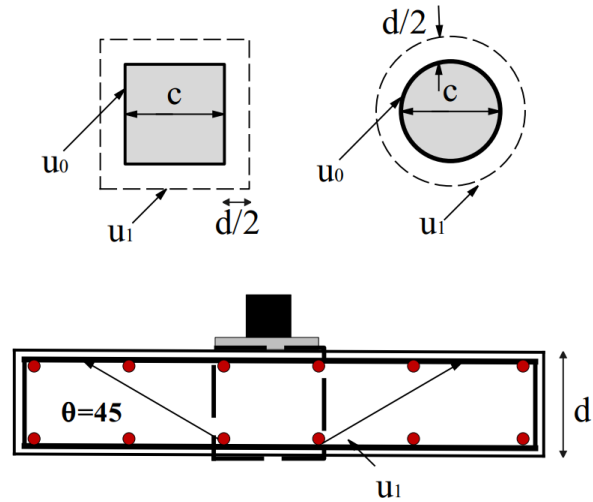

FIGURE 3: EC2-based boundary for SCS

$$V_{rc} = 0.18 * u_1 * d * \left(1 + \sqrt{\frac{200}{d}}\right) * \sqrt[3]{100 * f'_c * \rho s} \qquad (4)$$

and

$$\rho s = \sqrt{\rho_x \rho_y}, \text{ and } \rho_x, \rho_y, \qquad (5)$$

where $\rho_x$ and $\rho_y$ are x and y direction reinforcement ratio of the slab respectively. CDC is based on the truss analogy approach; however, CFP is based on the structural response of an arch-like frame at ULR as represented in FIGURE 4. The equations are given below,

$$w_{II,2} = W_C + (2\lambda_c d) \qquad (6)$$

$$\lambda^c = 2 - \left[\frac{100\rho_l f_y}{500}\right][1 + 0.01(f_c - 60)] \tag{7}$$

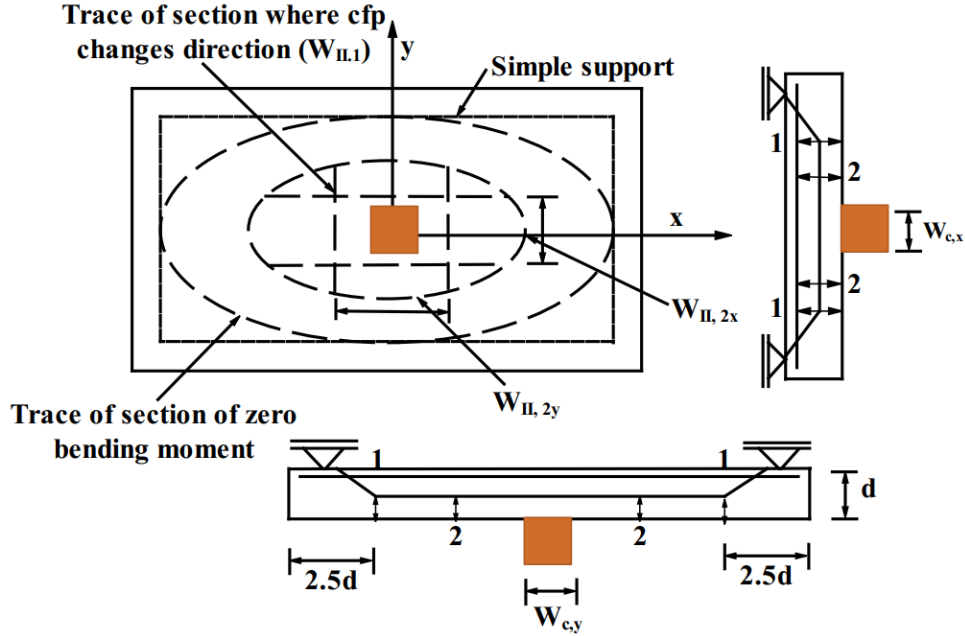

FIGURE 4: CFP-based punching failure assumption of SCS

## 3 Soft Computing Methods

Soft computing methods such as machine learning are an alternative to the analytical approach. This study uses three such techniques based on machine learning including Feed Forward Neural network (FNN), particle swarm optimization-based FNN (PSOFNN), and Bat algorithm-based FNN (BATFNN). The author's previous work (Ahmad, Arshid et al. 2021) included only ANN as a soft computing approach for determining the punching shear strength of flat slabs and since then the ACI code has also been revised. This study includes two more techniques of PSOFNN and BATFNN and an updated version of ACI 318-19 for calculating the shear strength of flat slabs.

### 3.1 Artificial Neural Network (ANN)

ANN is a powerful artificial intelligence tool developed to mimic the working of a human brain (Basheer and Hajmeer 2000). Neural networks have several unique features that enable them to be implemented in various fields of study. They can be used in data processing, image processing, prediction, and classification (Perera, Barchín et al. 2010). The ANN architecture consists of neurons and layers, where the layers are divided into input, hidden, and output layers (Shafabakhsh, Ani et al. 2015) that contain neurons. The input layer receives the data from the model, this data is transferred to the hidden layers. The neurons are linked together through weights which are multiplied by the neurons generated values which are then added with the bias. These weights are randomly initialized, and they are updated on each iteration to lessen the difference between the input and the predicted values for the later iterations. Multilayer Feedforward ANN (MLFNN) consists of an interconnected perceptron in which data flows from the input to the output layer. The number of layers in the network is the number of layers of the perceptron. A simple neural network consists of single input and output layer, this is called a one-layer FNN. Adding intermediate hidden layers to the network increases the complexity of the network. Hidden layers perform the computation on the data using activation functions that enable the output layer of the ANN to perform predictions (Demir 2008). The inputs and predictions are evaluated by the neural network for errors and these errors are propagated from the output node to the input node. The neural network keeps on reducing errors based on the provided conditions during the initialization of the network training. These conditions can be the number of iterations, the performance goal of the network, and maximum validation failures.

## 3.2 Particle Swarm Optimization

Particle Swarm Optimization is an intelligent optimization algorithm belonging to a class of nature-inspired algorithms called metaheuristics. It is based on the social behavior of animals like birds and fish. Fish and birds modify their movements for seeking food and avoiding predators (Poli, Kennedy et al. 2007). It is applied in various fields of science and engineering. PSO uses several particles to make a swarm that searches for the best solution in the search space. In PSO, every particle has three main parameters that are the velocity of the particle, particle position, and the previous best position of the particle (Premalatha and Natarajan 2009, Marini and Walczak 2015). With each successive iteration of the PSO, the particles try to converge to the best position $\vec{x}_i(t+1)$ by adjusting their position $\vec{x}_i(t)$ and velocity $\vec{v}_i(t)$ by keeping track of their experience as shown in **Error! Reference source not found.**.

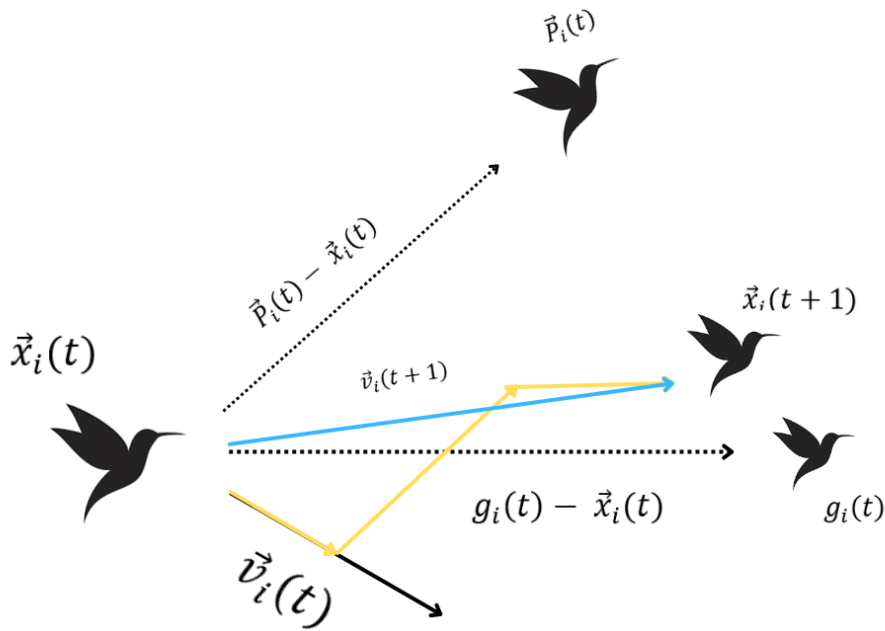

FIGURE 5: Adjustment of Velocity and Position in PSO algorithm

The particle in the swarm having the best personal value for fitness is taken as the global best. A summary of working of the PSO is creating a swarm population consisting of particles having a velocity and position, every particle has a memory of its best position that is termed as personal best and there exists a common best experience among members of the swarm known as global best. Every member updates their position based on their personal best and global best which helps them to converge (Nebro, Durillo et al. 2009). The working of the PSO algorithm is represented in a flowchart in FIGURE 6. In this study, the hyperparameters of the PSO are optimized using a grid search to identify the combination that results in optimal model performance. The hyperparameters of the PSO algorithm are swarm size, inertia weight, cognitive weight, social weight, maximum iterations performed for convergence, and velocity limits. The size of the swarm determines the number of particles employed for optimization. The potential of discovering a global optimum increase as the size of the swarm does, but it does become computationally expensive. The best value of the swarm size is based on the problem being solved as it depends on the complexity of problem, it dimensionality, size of search space thus it may vary.

The inertia weight, cognitive weight, and social weight are what govern the algorithm's behavior when it comes to exploration and exploitation. The influence of the particle's former velocity on its current movement is determined by the inertia weight. The inclination of the particle to travel toward its own optimal position is controlled by the cognitive weight. A higher value indicates that the particle's previous best position will have a bigger influence on the current movement. The tendency of the particle to move toward the swarm's optimal position on a global scale is controlled by the social weight. The inclination of the particle to go in the same direction as it did in the previous iteration will rise as inertia weight increases. Increasing cognitive weight can make the algorithm more aggressive in its exploration of the search space by increasing the tendency of the particle to advance towards its own personal best position. Increasing the value of all these three weights can make the algorithm converge faster but also increases the chances of getting stuck in a local optimum.

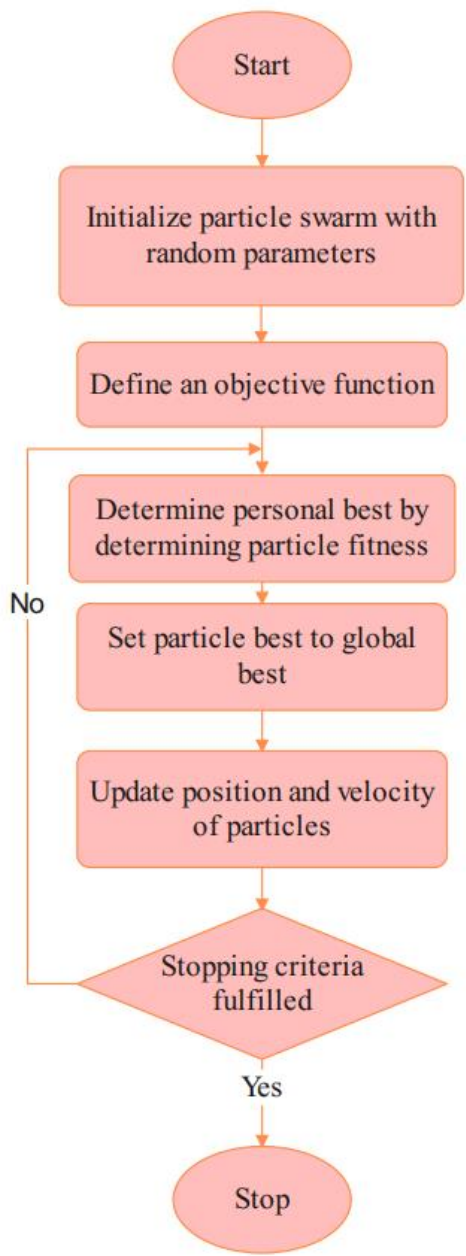

FIGURE 6: PSO Flowchart

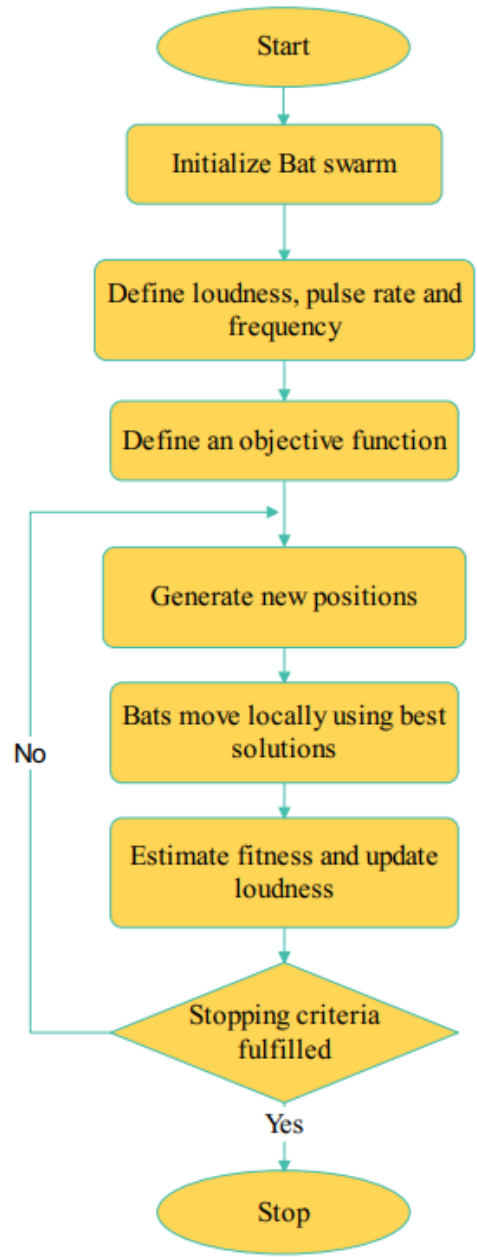

FIGURE 7: BA Flowchart

## 3.3 BAT Algorithm

The Bat Algorithm was developed by Xin-She Yang and combined the merits of existing optimization algorithms for developing the Bat algorithm (BA) (Yang and Gandomi 2012). BA is developed on the most important echolocation characteristic of bats that acts as a SONAR (Fister, Yang et al. 2014). The echolocation ability of bats enables them to not only search for prey but also differentiate between prey and obstacles in the path. Bats produce loud sound pulses and when these pulses bounce back from the objects in the path, bats listen to those echoes. Bats can vary the pulse frequency and the number of pulses produced per second, the pulse emission rate increases as the bats approach the prey. The loudness also varies with the pulse emissions, from loudest while searching for food to quieter as the prey is in proximity (Yang and Gandomi 2012, Fister Jr, Fister et al. 2013). Bats use the delay in the time of emissions, detection of echo, and variations in loudness to detect the orientation and distance of the prey, its type, and its speed. Bats have other sensitive senses like smell, and some even have good eyesight, but the BA is based on only echolocation. FIGURE 7 shows the Bat algorithm working from the initialization of bats to the definition of the objective function, to the optimization of the function and respective adjustment of the parameters involved until the stopping criteria, are fulfilled. The population size, maximum and minimum values of the search space, the frequency range, loudness, and pulse emission rate (alpha) are some of the parameters that affect the optimization process in Bat Algorithm. The careful selection of the hyperparameters for the Bat Algorithm (BAT) is crucial for the algorithm's performance in optimization procedures. These hyperparameters have a noticeable impact on the algorithm's behavior as it moves through the convergence, exploration, and exploitation stages. The strength of the algorithm's exploration and exploitation efforts is influenced by the "Loudness" parameter. The algorithm tends to explore new areas in the search space more thoroughly as "Loudness" is increased.

### 3.4 Database of Square Concrete Slab

A database comprised of 145 SCS samples is prepared for predicting the punching strength of slabs and the experimental setup used for the testing is shown in **Error! Reference source not found.**. The input parameters used are slab depth ($d_s$), column dimension ($c_s$), shear span ratio of flat slab ($a_v/d$), longitudinal reinforcement yield strength ($f_{yls}$), longitudinal reinforcement ratio ($p_{ls}$), ultimate load carrying capacity ($V_{us}$), and concrete compressive strength ($f_c$). The critical parameters are selected based on the physical models of CDCs. The influence of each parameter on the other parameter in the dataset is studied using the correlation heatmap, as shown in **Error! Reference source not found.**. This heatmap only includes the input and output variables present in the dataset. The parameter bo is the length of control parameter d/2 from the support, and bo = 4(c+d) for a square support, where c is width of column. The parameter a/d is not a parameter of database, therefore it is not used in the correlation Figure 10. The histogram in **Error! Reference source not found.** represents the correlation between the shear strength and various parameters of the flat slab dataset.

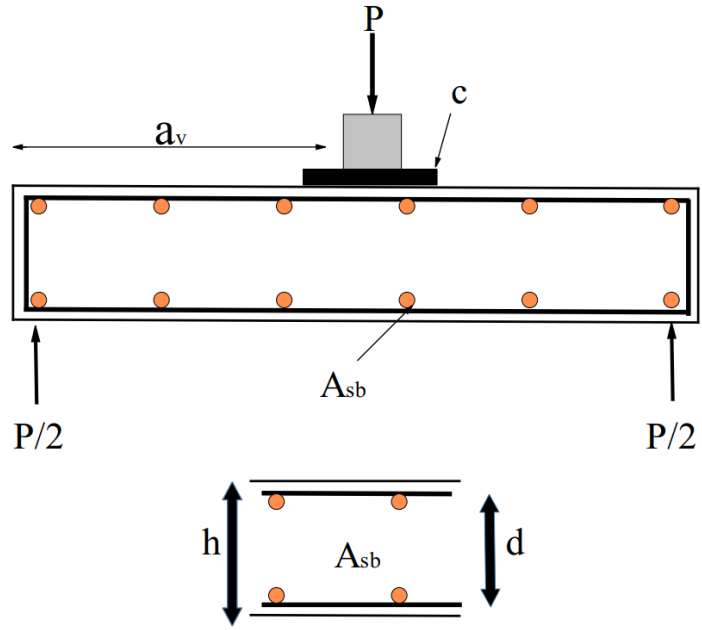

Figure 8: SCS experimental setup

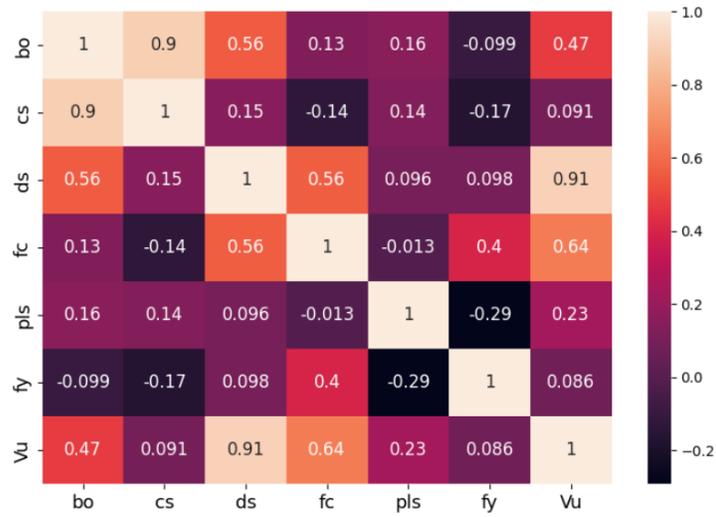

FIGURE 9: Correlation Heatmap for all parameters

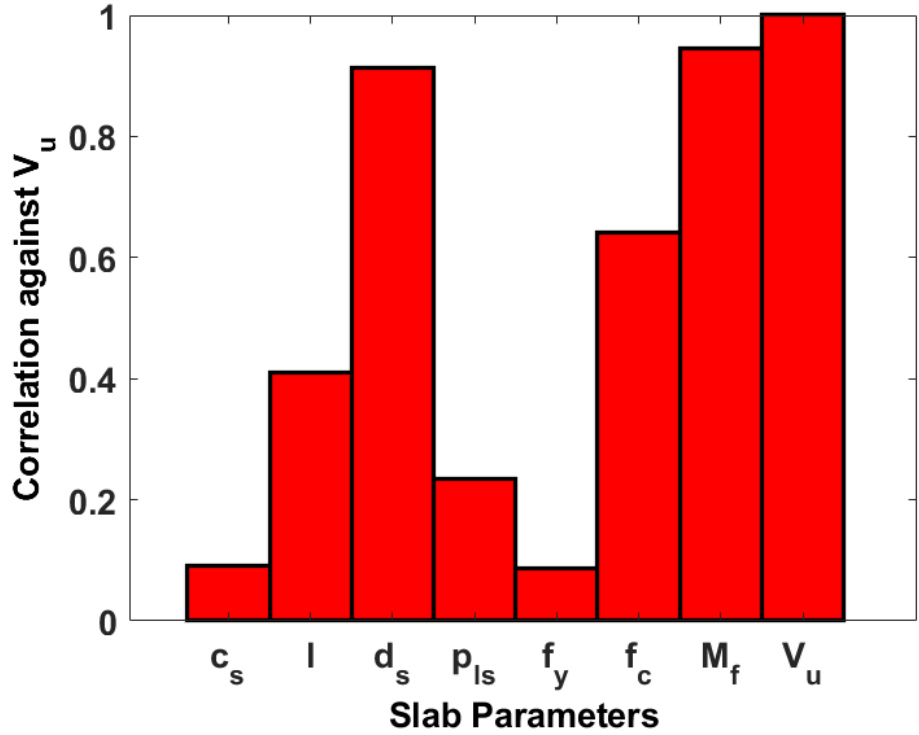

FIGURE 10: Correlation Histogram for all parameters

The heatmap shows that $d_s$ and $f_c$ have the greatest positive correlation with $V_u$, the dark region represents a low correlation, and the light region shows a higher correlation. The parametric slab data is summarized in Table **Error! Reference source not found.**, which includes the maximum, minimum values, the range of parametric values, their average value, and standard deviation.

Table 1: Parametric details of SCS database

| Parameter | Unit | Minimum | Maximum | Difference | Avg. | St. dev | COV |
|---|---|---|---|---|---|---|---|
| $d_s$ | (mm) | 64 | 275 | 211 | 122.32 | 44.82 | 0.37 |
| $c_s$ | (mm) | 54 | 600 | 546 | 206.34 | 87 | 0.42 |
| $a_v/d$ | | 4.5 | 14.02 | 9.52 | 7.81 | 2.4 | 0.31 |
| $\rho_l$ | (%) | 0.3 | 6.9 | 6.6 | 1.31 | 0.89 | 0.68 |
| $M_{fs}$ | (kN-mm) | 39000 | 1951000 | 1912000 | 252655 | 292121 | 1.16 |
| $f_{yls}$ | (MPa) | 294 | 749 | 455 | 496.88 | 117.68 | 0.24 |
| $f_{cs}$ | (MPa) | 9.52 | 118.7 | 109.18 | 41.3 | 24.85 | 0.6 |
| $V_{us}$ | (KN) | 105 | 2450 | 2345 | 458.7 | 436.88 | 0.95 |

## 3.5 Normalization of Data

Normalizing the database increases the efficiency of the ANN and the data can be denormalized at the end. Normalization helps in the conversion of data to unitless values that aid in making a visual correlation between data samples easier. Additionally, ANN shows low learning rates for unnormalized values (Krogh and Vedelsby 1994, Rafiq, Bugmann et al. 2001), therefore normalizing the values between suitable upper and lower boundary values is a better practice. The normalization can be done by either using the built-in functions of the programming package being used or manually. Doing it manually provides better control over the normalization process, like using different upper and lower limits. For this study, the data is normalized between 0 and 1 using equation (8).

$$X = \frac{X_o - X_{min}}{X_{max} - X_{min}} \tag{8}$$

$X_{max}$ and $X_{min}$ are the respective maximum and minimum values of the variable and $X_o$ is the current value that is normalized to $X$.

### 3.6 Prediction Models

Several error metrics like Mean Squared Error (MSE), Mean Absolute Error (MAE), and Pearson's correlation coefficient are used for selecting the ANN model (Kotsovou, Ahmad et al. 2020). The ANN model with the least values for MSE, MAE, and highest value for the R is used to train and test the dataset. ANN used for the database training and testing is MFNN, coded in MATLAB (Ahmad and Raza 2020). The hybrid models of PSOFNN and BATFNN use an objective function for optimizing the research problem. The objective function is optimized to maximize or minimize the fitness value of the hybrid model. For this study, MSE is used as an objective function in the hybrid models. The analytical expression for the MSE is given by the equation,

$$MSE = \sum_{i=1}^{n} \frac{(Y - \hat{Y})^2}{n} \tag{9}$$

where Y denotes the original target values, $\hat{Y}$ represents the predicted values of the target variable and *n* is the total number of available samples.

The hybrid models and the ANN model resulting in the lowest value for the MSE, MAE, and the highest value for R is utilized for training and testing of data. The parameters for the hybrid models are selected based on the output values of error metrics. The models are run iteratively using different values for each parameter, for determining the optimum values of the parameters in the PSOFNN and BATFNN. Important parameters for the PSO algorithm are swarm population, inertia weight, personal and global learning coefficient, no. of iterations, and upper and lower bounds. For the Bat algorithm, the critical parameters are no. of bats, iterations, and constant for loudness and rate of pulse emission. The number of layers in FNN, no. of particles and the no. of iterations in PSOFNN and BATFNN used for each dataset are given in the **Error! Reference source not found.**. The variation of the parameters in the hybrid model can help in the training and testing of the data. One might assume that increasing the swarm population or the number of iterations for the optimization algorithm will increase the efficiency of the hybrid models, but this does not work always. The increase in the parameters does not always prove to be beneficial for the optimization of the objective function. Therefore, the real challenge is to vary the parametric values until you get the best combination. Sometimes, the optimization efficiency will be the highest at the beginning which will optimize the results by more than 90% in the first few epochs, but it will not show much change as the iterations proceed.

Table 2: Parametric combination for SCS dataset for ML models

| No. | Dataset name | ANN layers | PSOFNN members | PSOFNN iterations | BATFNN members | BATFNN iterations | Parametric combination |
|---|---|---|---|---|---|---|---|
| 1. | SCS=1 | 14 | 50 | 300 | 30 | 300 | $\rho_{ls}$, $f_{yls}$, $f_c$, $c_s$, $d_s$, $a_v/d_s$ |
| 2. | SCS=2 | 12 | 50 | 300 | 30 | 300 | $M_{fs}$, $f_c$, $c_s$, $d_s$, $a_v/d_s$ |
| 3. | SCS=3 | 8 | 50 | 300 | 30 | 300 | $M_{fs}/f_c bd_s^2$, $c_s/d_s$, $a_v/d_s$, |
| 4. | SCS=4 | 10 | 30 | 300 | 30 | 300 | $\rho_{ls}$, $f_c/f_{yls}$, $c_s/d_s$, $a_v/d_s$, |
| 5. | SCS=5 | 10 | 50 | 300 | 25 | 300 | $M_{fs}/bd_s^2$, $f_c$, $d_s$, $a_v/d_s$ |
| 6. | SCS=6 | 10 | 50 | 300 | 25 | 300 | $M_{fs}/f_c bd_s^2$, $d_s$, $c_s/d_s$, $a_v/d_s$, |
| 7. | SCS=7 | 12 | 30 | 300 | 25 | 300 | $M_{fs}/f_c bd_s^2$, $f_c$, $d_s$, $c_s/d_s$, $a_v/d_s$ |

## 4 ANN, PSOFNN, and BATFNN models for SCS

ANN, PSOFNN, and BATFNN are run on seven different parameter sets for the SCS data. The output of the PSOFNN, BATFNN and ANN models ran on these seven datasets. The original SCS data is used to develop these seven combinations of data, which enables us to use seven ML models on them, to determine the effect of the difference of the dataset on the target value. Judging the ANN and hybrid models only based on predicted and experimental value plots are not preferred as the error metrics are vital in deciding the best working model. The PSOFNN predictions for $V_u$ represent that the models for SCS=1 and SCS=2 are better than the other five models. The MSE and MAE values for these two are the lowest which are 0.0275%, 1.214% for SCS=1, and 0.0316%, 1.481% for SCS=2, as shown in FIGURE 13 and **Error! Reference source not found.**. Based on the predicted values of the $V_u$, error metrics and the R-value, PSOFNN for SCS=1 is the best model having an R-value of 99.23%. FIGURE 11 and FIGURE 12 illustrated the results for the ANN models, and FIGURE 13 and **Error! Reference source not found.** demonstrate result for PSOFNN and FIGURE 15 and FIGURE 16 for BATFNN.

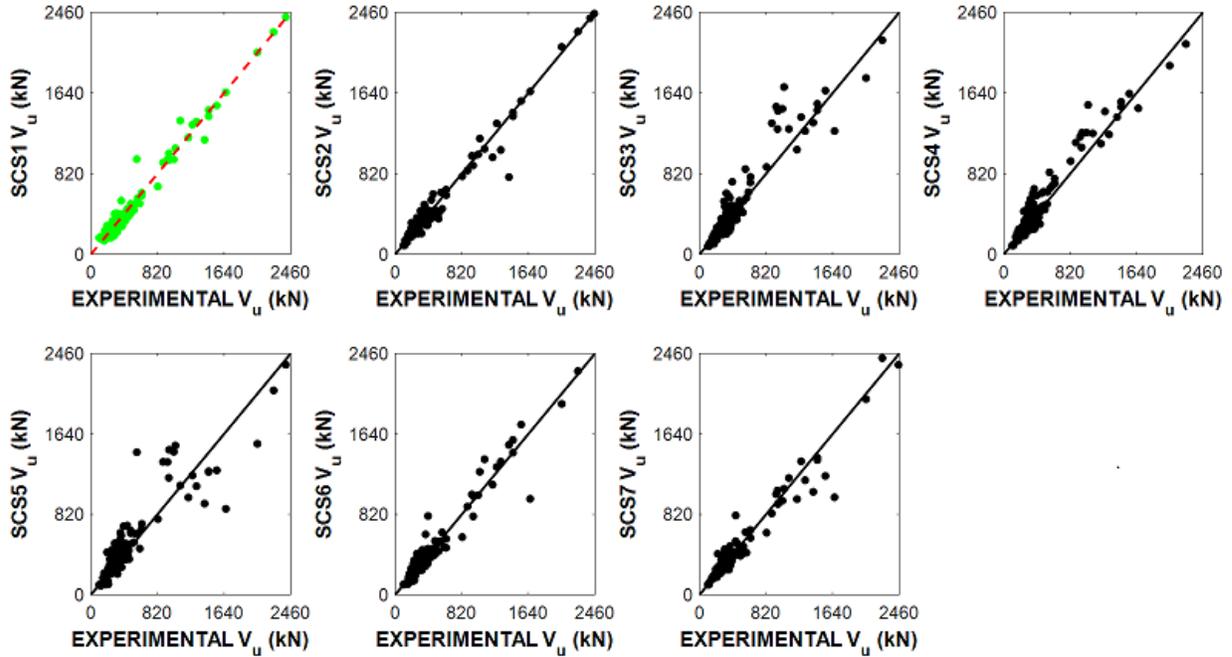

FIGURE 11: ANN-based predictions for shear strength

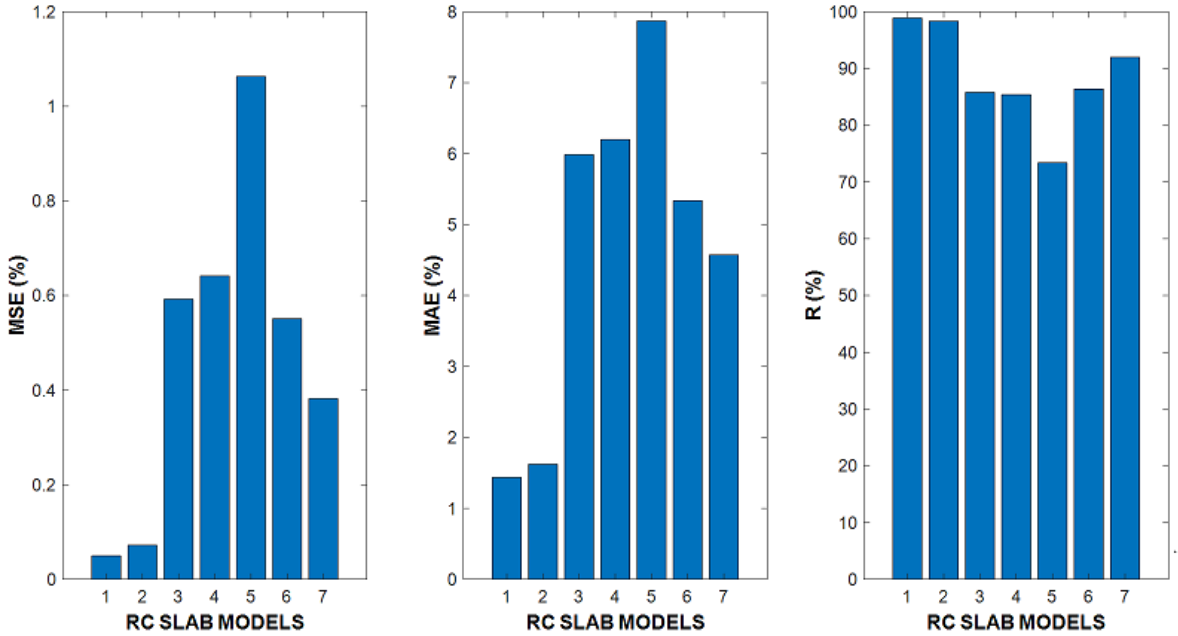

FIGURE 12: Error metrics for ANN

A comparison of the error metrics predicted values and the accuracy of the PSOFNN, BATFNN, and ANN models shows that PSOFNN has the best-predicted results with the lowest error. This can also be interpreted as the best optimization performance of the PSO on the MSE function. The performance is determined by the convergence of the algorithm around the desired result that in our case is the reduction of the MSE. PSOFNN has the lowest MSE values for all seven models as compared to BATFNN and ANN. The value of R for PSOFNN is either higher than ANN or comparable but never less for any of the seven datasets. ANN is the second-best model out of the three ML models, while the BATFNN results show that the BAT algorithm does not optimize the objective function well as can be seen in FIGURE 15. The working of the hybrid models can be improved by finding appropriate parametric values for the algorithm. Moreover, the performance also varies with the objective function used for optimization. FIGURE 13 and **Error! Reference source not found.** demonstrate that PSOFNN performed better than BATFNN for the MSE objective function for all seven combinations of the dataset.

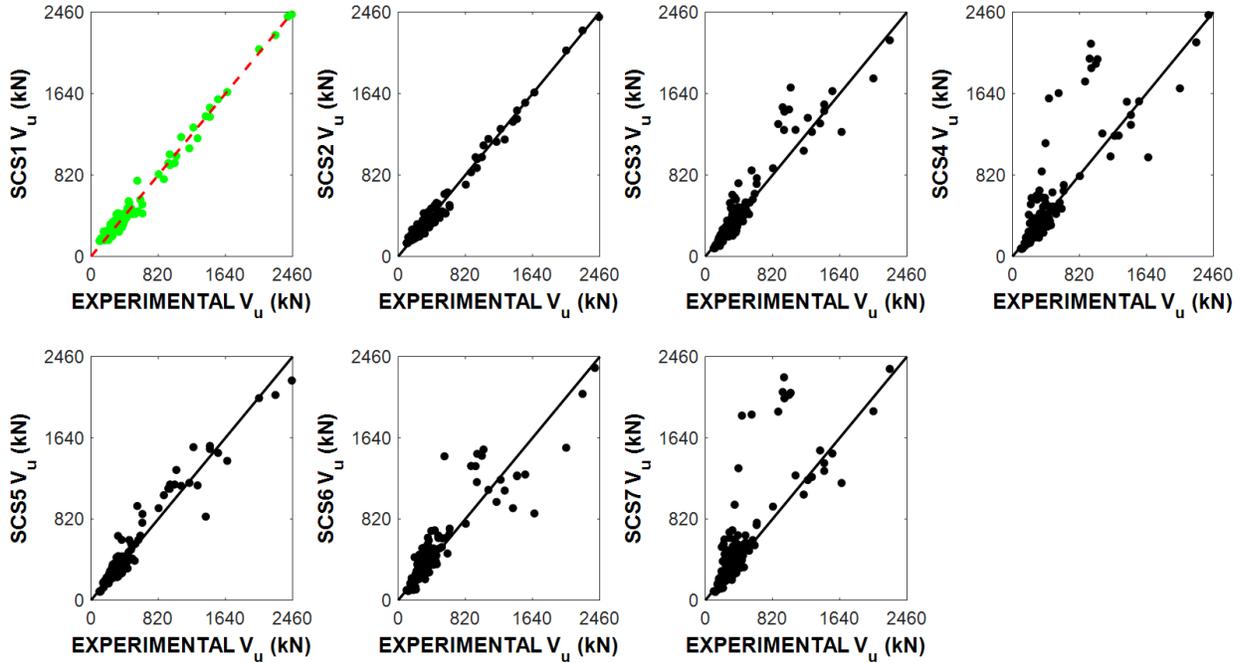

FIGURE 13: PSOFNN predictions of shear strength

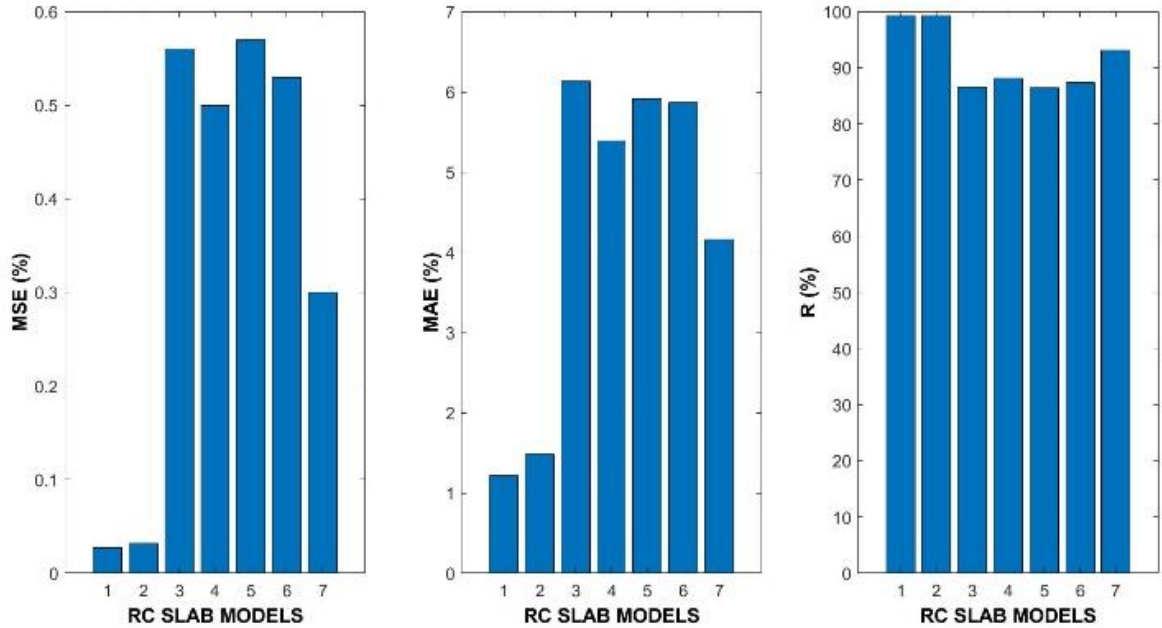

FIGURE 14: Error metrics for PSOFNN

ANN prediction ability varied significantly with the variation in the number of input variables as shown in FIGURE 12, the MSE, MAE, and R values are the highest for the SCS=5. The SCS parametric combination of the 3$^{rd}$, 4$^{th}$, 5$^{th,}$ and 6$^{th}$ dataset has either 4 or less than 4 inputs. ANN performed better for the SCS dataset with 5 or more input variables. ANN and BATFNN showed a sudden performance drop for at least one of the SCS combinations, but PSOFNN performed equally well for all datasets with a similar number of inputs as shown in FIGURE 16.

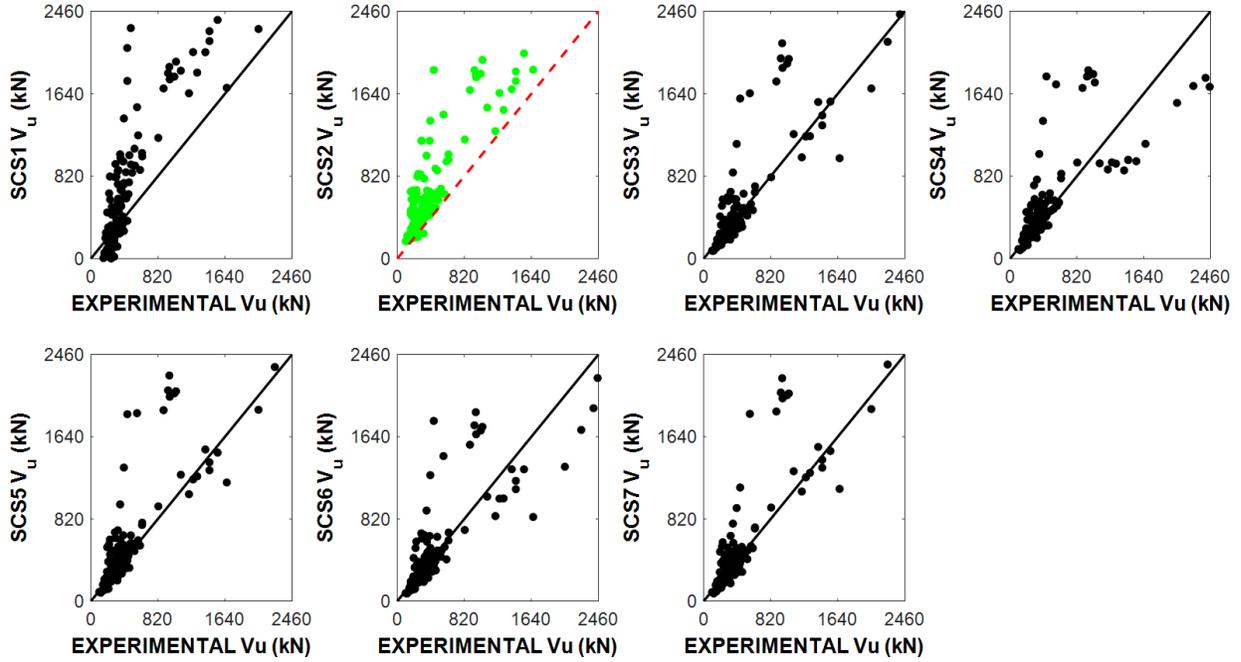

FIGURE 15: BATFNN predictions for shear strength

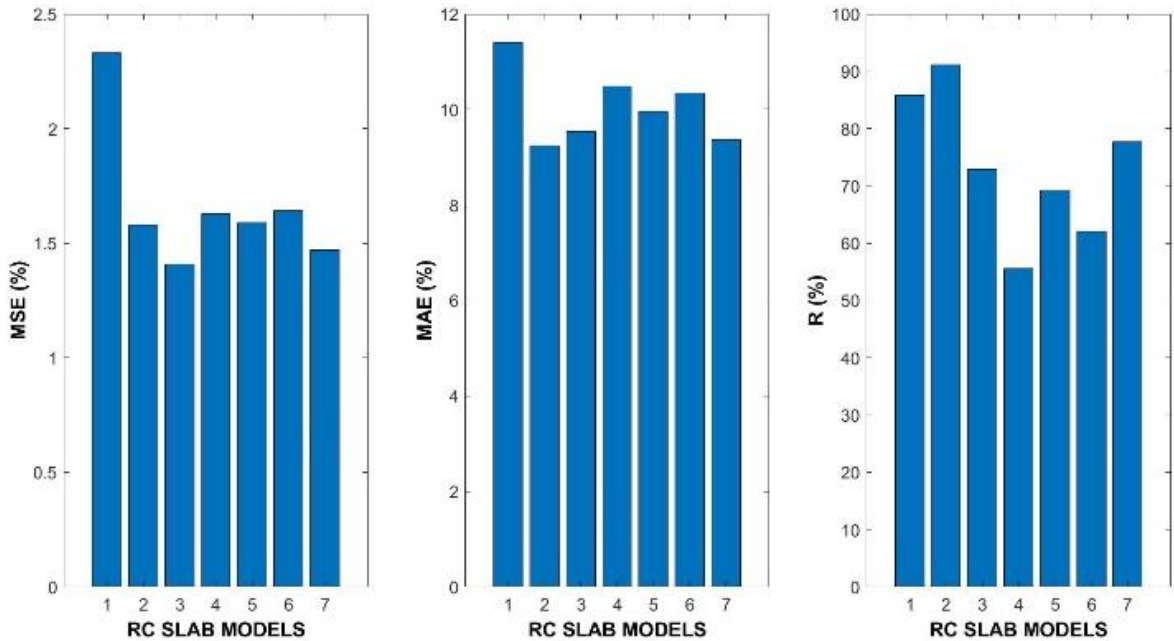

FIGURE 16: Error metrics for BATFNN

## 5 Comparative Studies of ML Models, CFP, and CDCs

In this section, a comprehensive analysis of the results obtained from six different approaches used for the shear strength at slab-column connections. These six approaches includes use of current deign codes of ACI 318-19 and Eurocode 2, Compressive Force Path method (CFP) method, and three neural network based models including Feed Forward Neural Network (FNN), Particle Swarm Optimized FNN (PSOFNN) and Bat algorithm based FNN (BatFNN). The performance of each model is assessed using three key metrics including Mean Square Error (MSE), Mean Absolute Error (MAE), and coefficient of determination ($R^2$). These evaluation metrics provide information about the accuracy, precision and fitness of models compared to the experimental values in the dataset. A thorough analysis show that PSOFNN is the most reliable model for shear strength prediction of flat slabs and it outperforms all other models in every evaluation metric and variation of dataset. The PSOFNN provided the lowest values for the error metrics and highest value for $R^2$. It should be noted that the design codes of ACI 318-19 and EC 2 underestimate the punching shear capacity. Eurocode 2 (EC 2) demonstrated a higher accuracy in representing the experimental results than the ACI 318-19. The error metrics of the EC 2 were also lesser than for the ACI 318-19. The best prediction results of each model are shown in FIGURE 17 and the error metrics including MSE, MAE $R^2$ for each model are shown in FIGURE 18. From these results it becomes evident that PSOFNN is the best model for prediction of punching shear strength of flat slabs with $R^2$ value of 99.37%, MSE of 0.0275% and MAE of 1.214%. The value of MSE and MAE for EC 2 are less than ACI 318-19 and similarly the $R^2$ value for EC 2 is significantly greater than for the ACI 318-19, indicating greater accuracy in determining punching shear strength of flat slabs.

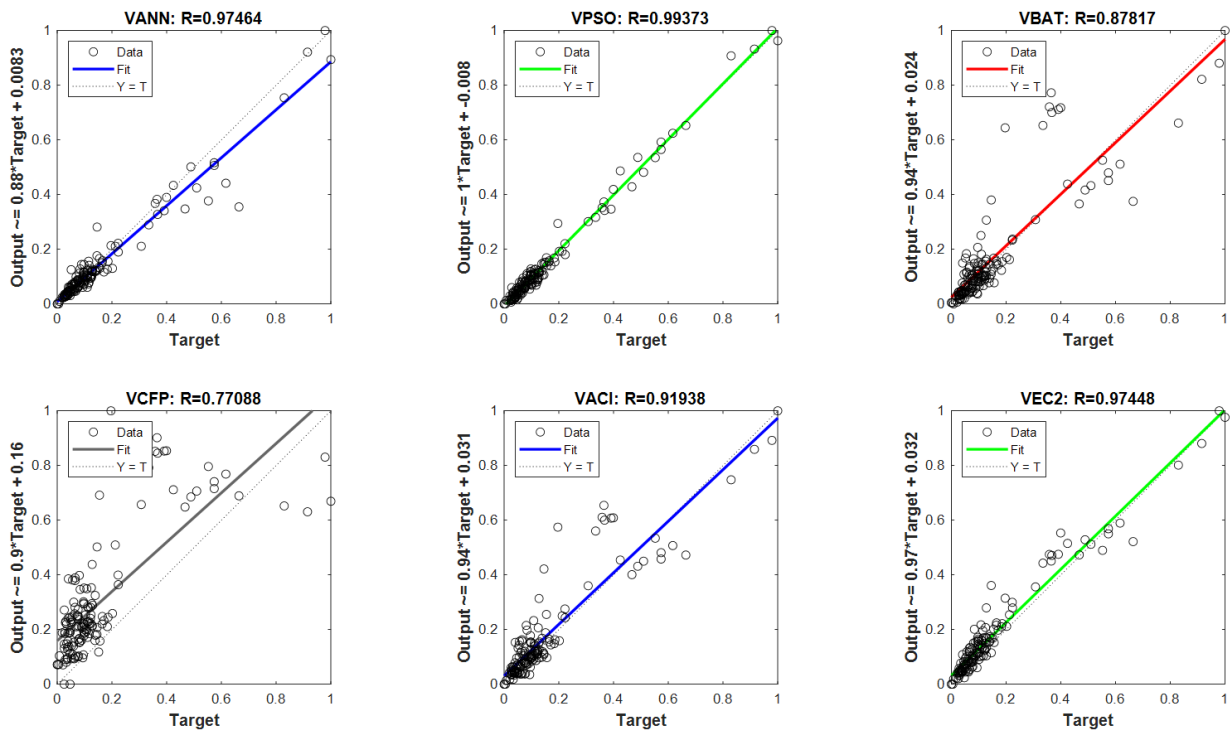

FIGURE 17: Value of R for Six Models

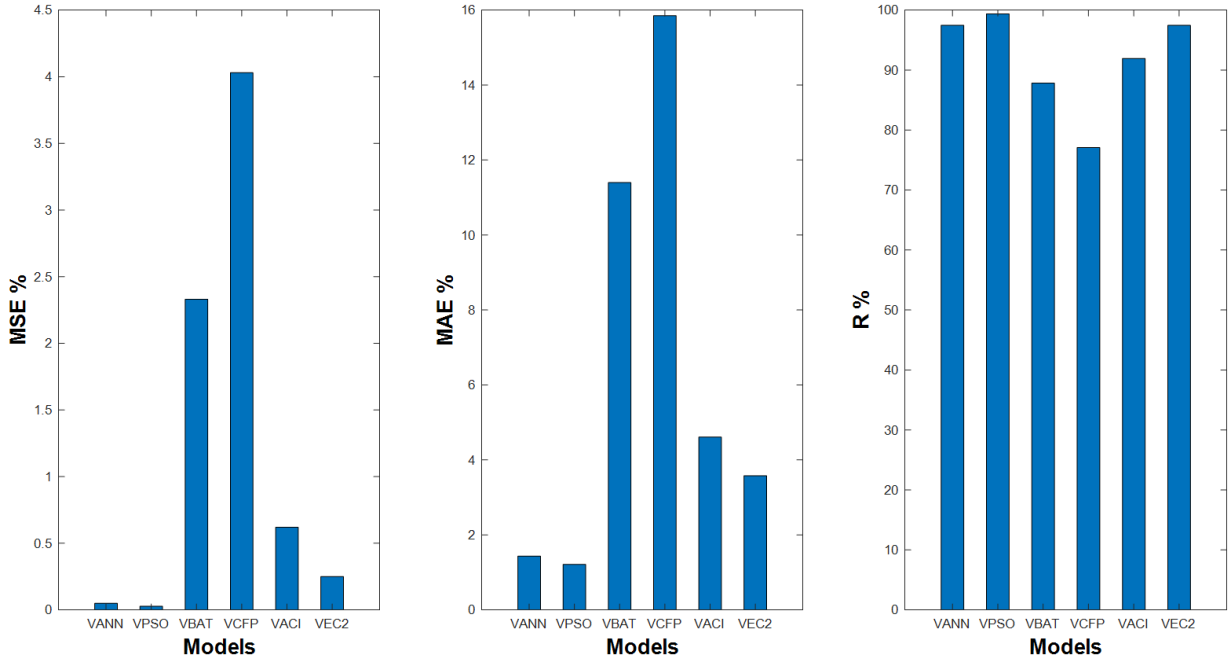

FIGURE 18: MSE, MAE and R for Six Models

The normal distribution of the shear strength ratios for all models is shown in FIGURE 19. The shear strength ratio plot for the machine learning models of FNN, PSOFNN and BATFNN are distributed uniformly around 1 and width of their distribution curve is narrow that suggests that the shear strength ratios for these models are less dispersed and are closely clustered around the mean indicating good prediction results. In contrast, a wider curve or higher standard deviation of the shear strength ratio results for the other models including CFP, ACI 318-19 and EC 2 indicate greater dispersion of values from the mean. The range of ratio of experimental and predicted shear strength values for all models is given in FIGURE 20. This analysis suggests that PSOFNN stands out as the best model overall while EC 2 demonstrates a superior performance in the domain of design codes.

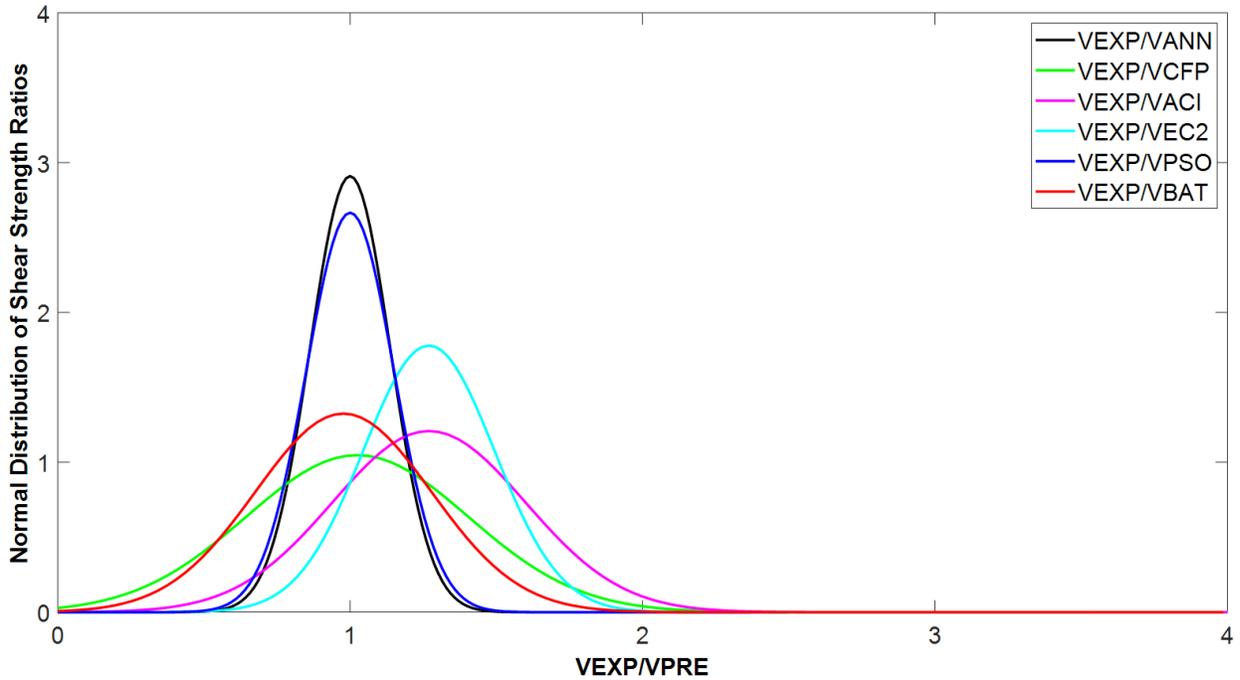

FIGURE 19: Normal Distribution of Shear strength ratios

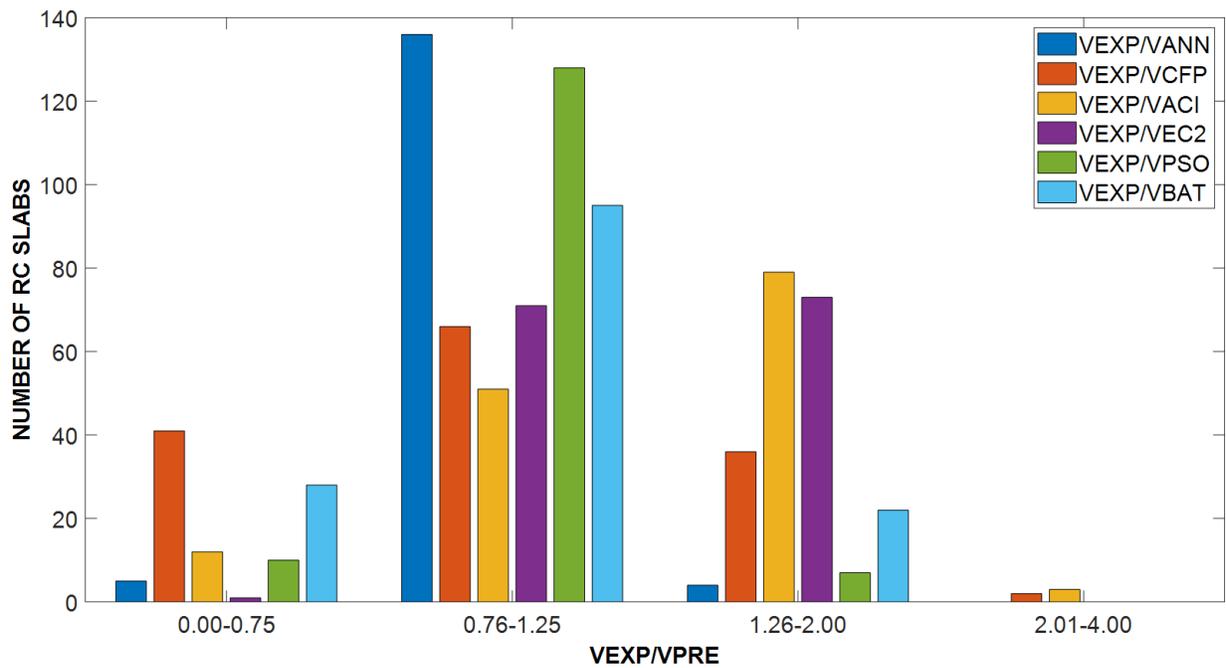

FIGURE 20: Range limit for SCS ratios

## 6 Analysis of Results for Different Parameters

The effect of flat slab parameters on the prediction results of the CFP method, CDCs, ANN, PSOFNN, and BATFNN is discussed here. The parameters used for the parametric studies are slab depth ($d_s$), column dimension ($c_s$), shear span ratio of flat slab ($a_v/d$), longitudinal reinforcement yield strength ($f_{yls}$), longitudinal reinforcement ratio ($p_{ls}$), ultimate load carrying capacity ($V_{us}$), and concrete compressive strength ($f_{cs}$). The parametric study is done to determine the influence of these parameters on the ratio of $V_{EXP}/V_{PRED}$.

### 6.1 Effective Depth of Slab (ds)

The plots for the prediction of SCS punching failure response as illustrated by the FIGURE 21 show that the effective slab depth ($d_s$) is one of the most influential geometric features for SCS punching failure. The results of the parametric comparative analysis for PSOFNN are better than the ANN, BATFNN, CDCs, and CFP. The results of PSOFNN and ANN are closer to the experimental value of ratios and the unity line is used as a reference for determining the fit of the ratios. The plots of the CDCs ratio in FIGURE 21 against $d_s$ show that most of the values are above the unity line, indicating that CDCs result in values lesser than actual experimental values. The scatter of points for the PSOFNN is closer to the unity line, as shown in FIGURE 21, indicating that PSOFNN predicted values are closer to the experimental values. The results can be used to further analyze the reason which results in lesser values for shear strength than actual. The values of $d_s$ in the range of 75mm to 135mm showed the highest deviation from the unity line. This requires revising the CDCs as they are underestimating the values of the punching shear, this means that the actual punching shear at the column-slab connection is larger than the values calculated using CDCs, which can lead to punching shear failure.

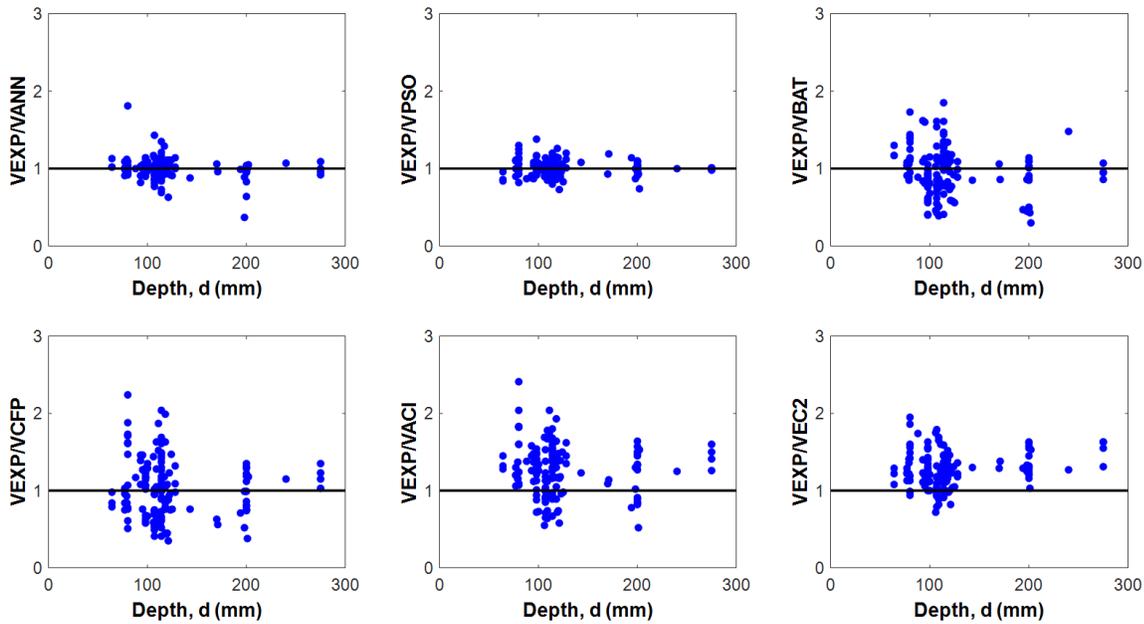

FIGURE 21: Shear Strength ratio against depth

### 6.2 Shear Span ratio ($a_v/d$)

The effect of shear punching failure in flat slabs is studied for the experimental and predicted values ratio against the shear span ratio of flat slab. The scatter of ratios on the unity line in FIGURE 22 shows that the values are closer to unity line for the PSOFNN and ANN, and the scatter is also uniform on the top and below of unity line for ANN, PSOFNN, BATFNN and CFP. Although the values of the ratio are scattered more widely for the CFP but there is no clear biasness towards underestimation or overestimation of values. ACI and EC2 codes show the highest biasness of values of ratios towards lesser values than original, as shown in FIGURE 22.

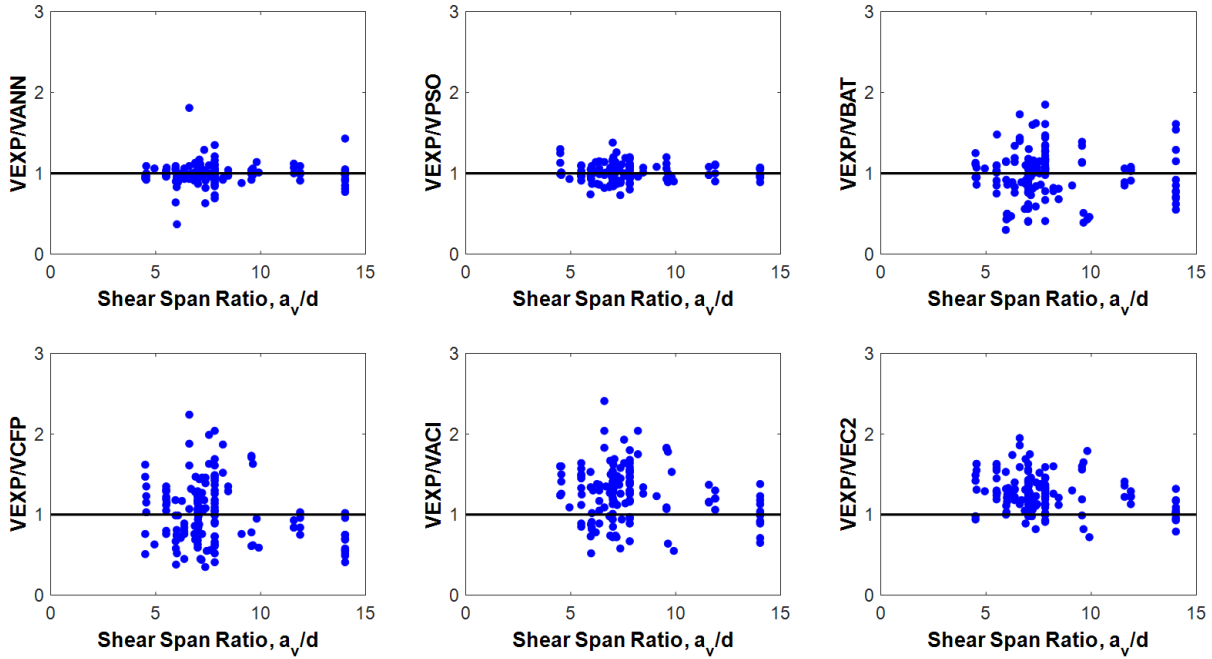

FIGURE 22: shear strength ratio against span ratio

## 6.3 Compressive Strength ($f_{cs}$)

For the SCS, punching failure can be predicted more precisely by using compressive strength ($f_{cs}$) as illustrated in FIGURE 23. The results of the PSOFNN and ANN are close to the experimental results, therefore the scatter of the points is close to the unity line, while the scatter of VEXP/VACI and VEXP/VEC2 against the unity line show that ACI and EC2 underestimate the values. The ideal case requires these values to lie on the unity line but the presence of most points above the unity line is proof that values calculated by the CDCs are less than the actual experimental values. The highest deviation resulted by the values in the range of 20MPa to 40 MPa for $f_{cs}$. The scatter of points in the CFP and BATFNN plots in FIGURE 23 show that the values for these two are roughly uniformly distributed around the unity line. This indicates the non-biasness of these two approaches towards overestimation or underestimation of the values.

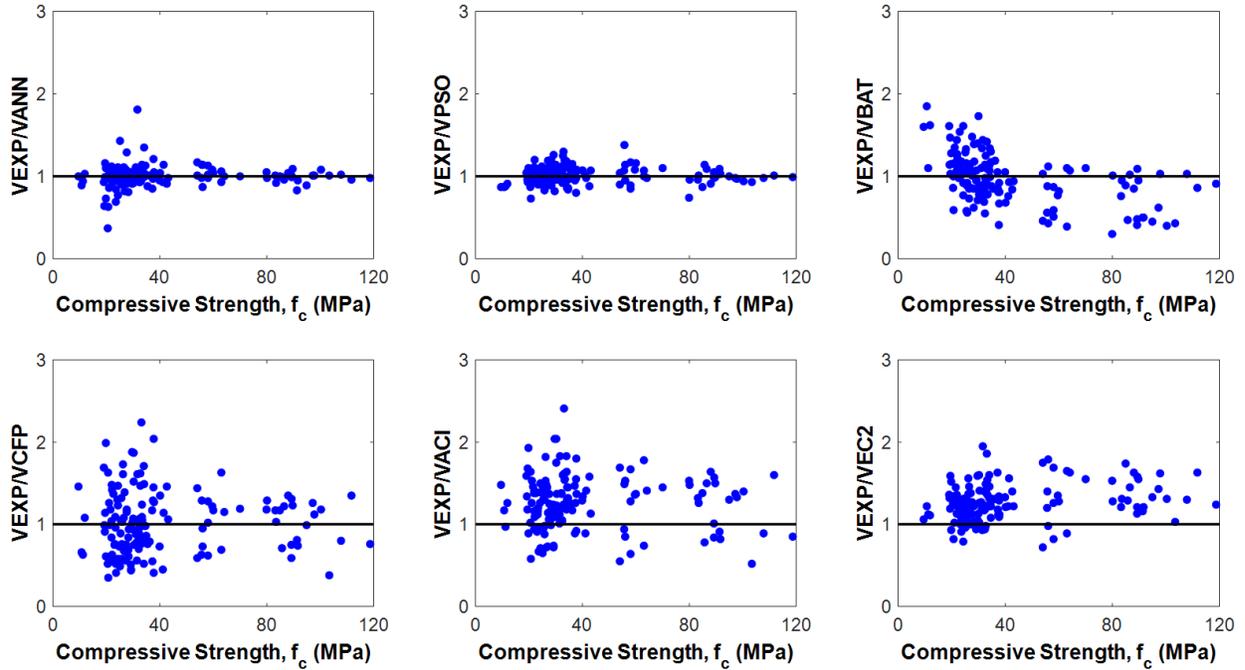

FIGURE 23: Shear strength ratio against Compressive Strength

## 6.4 Longitudinal Reinforcement ratio ($p_{ls}$)

The flat slab samples having the reinforcement ratio value between 1.5% and 2.5% show the most deviation from unity line, as shown in **Error! Reference source not found.**. The reinforcement ratio is one of the two most important material property for the prediction of punching shear failure response of flat slab. The scatter of the points is closer to the unity line for the PSOFNN and ANN which shows that the PSOFNN and ANN have performed better predictions than the rest of the 4 models. The BATFNN has performed less accurate predictions than PSOFNN and ANN but the results are less scattered than the CFP and ACI codes. In addition, the ANN, PSOFNN and BATFNN have shown no biasness in predicting the punching shear strength either towards underestimation or overestimation.

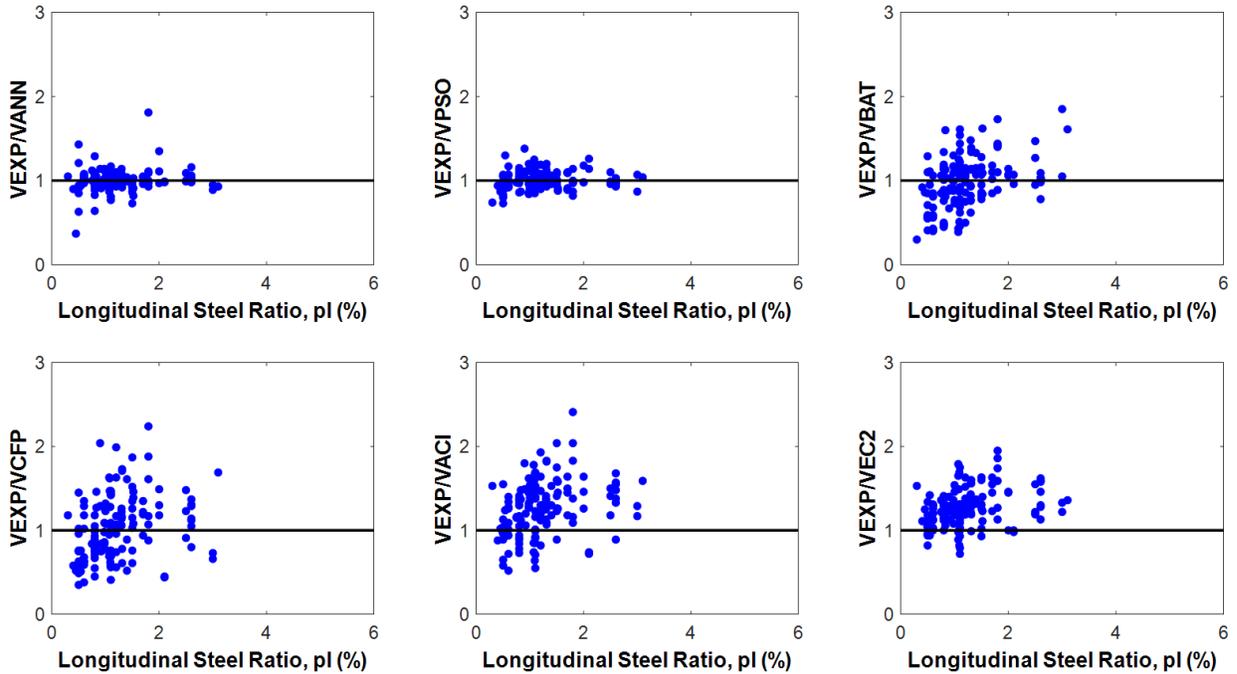

FIGURE 24: Shear strength ratio against longitudinal steel ratio

## 7 Using NLFEA for RC Flat Slab Assessment

The behavior of concrete is non-linear and in ABAQUS, this behavior is studied using the concrete-damaged plasticity and smeared crack model. The compression and tension region in concrete or the biaxial tension region result in cracking of concrete at the failure in case of smeared crack concrete modeling (Hu and Schnobrich 1990). The smeared crack modeling is used for representing concrete crushing due to compression or cracking because of tension respectively. Under uniaxial tensile loading, concrete behaves elastically for as long as the tensile loading is in the range of 7.5% to 11% of final compressive stress, after this the concrete cracking begins. It is assumed that cracking will occur when the stress in concrete approaches the failure surface known as the "crack detection surface". The cracking damages the structure and the smeared cracking model also considers the cracking as damage. The cracks under compressive stress are considered to be completely closed because these types of cracks do not show any strain.

### 7.1 Concrete Damaged Plasticity Model

In ABAQUS, the behavior of concrete material in the inelastic range is defined using concrete-damaged plasticity model. This model can be used for plain concrete, but its main purpose is to analyze reinforced concrete. Concrete subjected to low confining pressures or monotonic and cycling loading can be analyzed using this model. The model considers two main failure mechanisms of concrete, one by crushing concrete and another by tensile cracking. Under the action of uniaxial tensile stresses, the stress and strain follow a linear elastic relationship till the failure occurs. As the tensile stresses increase beyond the failure stress, strain localization occurs in the concrete. When the uniaxial compression is applied, the response is linear till the initial yield point leading to stress hardening state in the plastic region that causes strain softening as the stresses increase beyond ultimate stress. The behavior of concrete is brittle under low confining pressures and the CDP model represents this failure using stress and plastic strain.

The non-linear behavior of concrete slabs is studied using ABAQUS. A quarter portion of the slab is modeled and analyzed in ABAQUS to reduce the analysis time required for the analysis of the full slab. The continuity condition is fulfilled using the boundary condition for the quarter portion of the slab. Several important material and geometric parameters are used to calibrate the slab model including the size of the mesh, dilation angle, eccentricity, viscosity, uniaxial to biaxial stress ratio, and various element types of concrete and steel. Load deflection curves are plotted with MATLAB using the results obtained from the load-displacement analysis of the slab.

The elastic stiffness of the sample undergoes degradation when it is unloaded in the strain-softening region of stress and strain curves. The stress-strain relationship under the action of uniaxial compression and tension is calculated using the following two equations.

$$\sigma_c = (1 - d_c)E_o\left(\varepsilon_c^{\sim in} - \varepsilon_c^{\sim pl}\right) \tag{10}$$

$$\sigma_t = (1 - d_t)E_o(\varepsilon_t^{\sim in} - \varepsilon_t^{\sim pl}) \tag{11}$$

The damage in the elastic stiffness is represented by two variables *dt* and *dc*, whose value depends on the temperature, plastic strain, and field parameters involved. The value of these damage variables varies from zero to one, representing no damage and complete strength loss respectively. The other parameters involved in the equations are the initial elastic stiffness of the material ($E_o$), compressive strain ($\varepsilon_c$), tensile strain ($\varepsilon_t$) and the superscripts ~in and ~pl represent inelastic strain and plastic strain respectively. The concrete damaged plasticity model is used to model the plastic behavior of concrete using various parameters involving dilation angle, viscosity, load eccentricity, ratio of uniaxial to biaxial stress and shape factor of yielding surface as mentioned in **Error! Reference source not found.**. This data has been used to perform FEA of some selected cases of flat slab that are mentioned in **Error! Reference source not found.**. The tension stiffening simulation is defined in the slab model to represent the strain-softening behavior of cracked concrete. This tension stiffening is specified in one of two ways either by using a fracture energy cracking criterion or the stress-strain relationship of concrete after failure.

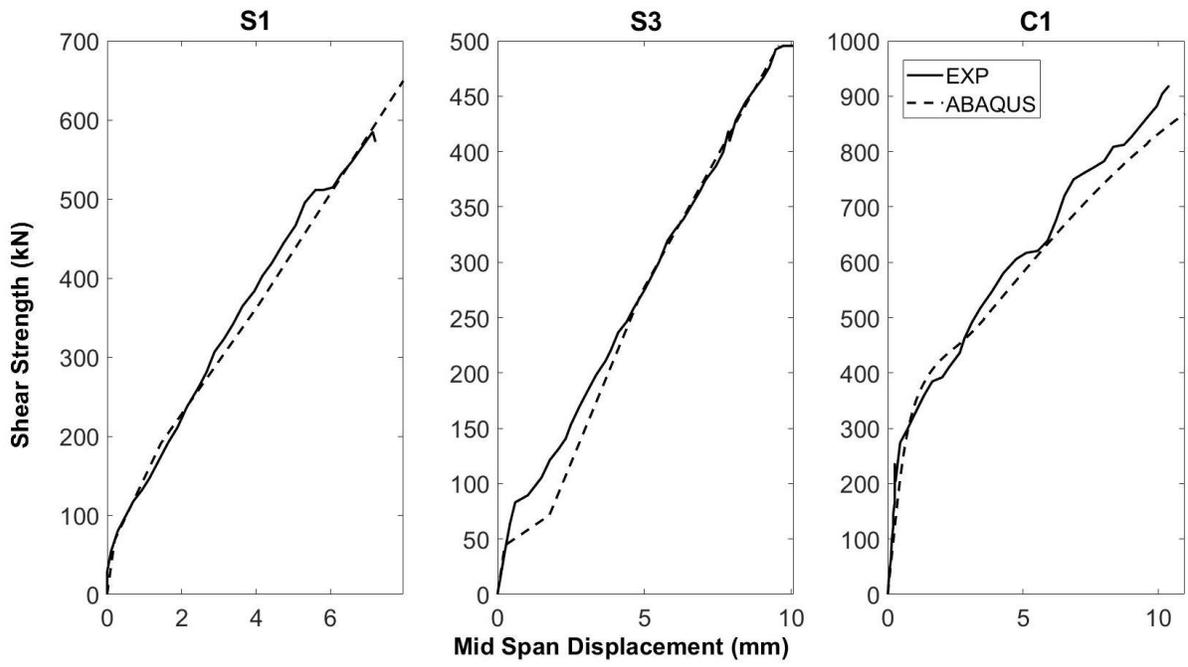

(a)

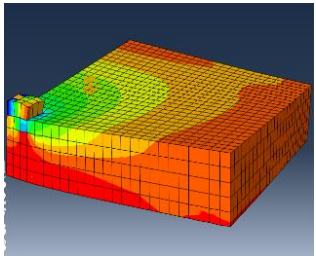 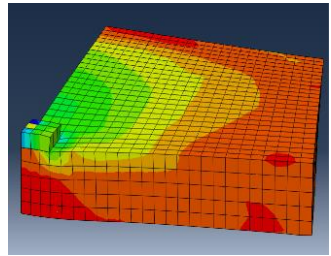 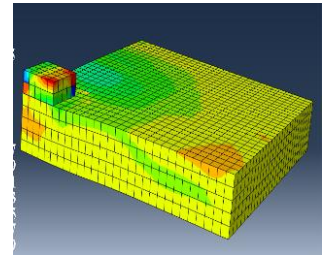

(b)

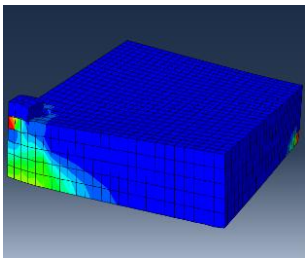 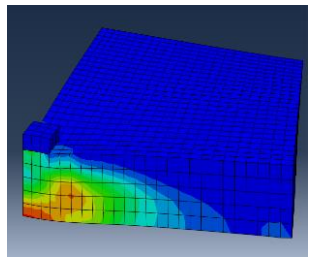 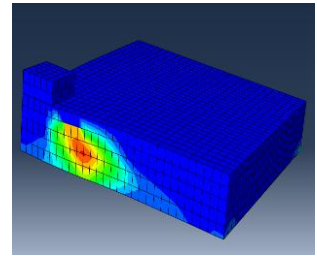

(c)
29

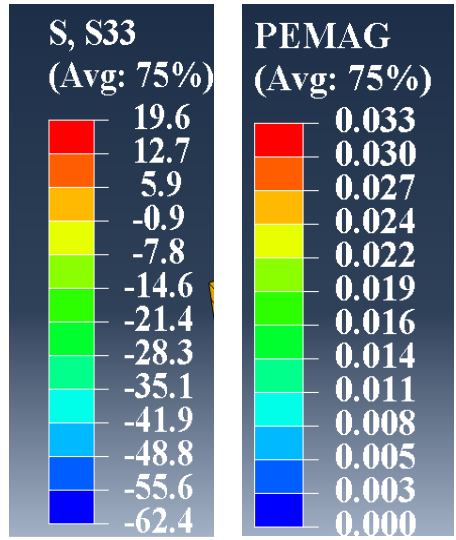

(d)

FIGURE 25: (a) LD Curve, (b) S33 (c) PEMAG (d) LEGEND for SCS (Ahmad, Arshid et al. 2021)

In ABAQUS, the steel bars and concrete are considered to be perfectly bonded. ABAQUS uses truss elements to represent the steel bars having two nodes with three translations at each. The material properties used for steel bars are yield strength, modulus of elasticity, and Poisson's ratio with respective values of 420 MPa, 200 GPa, and 0.3. The results of the ML models are verified using the analysis results from ABAQUS. The load-deflection curves from ABAQUS are in good relationship with the curves from experimental data, as shown in FIGURE 25(a).

Table 3: Selected samples for SCS

| Source | Name | $d_s$ (mm) | $c_s$ (mm) | $a_{vs}/d_s$ | $f_{cs}$ (MPa) | $f_{ys}$ (MPa) | $\rho_{ls}$ (%) |
|---|---|---|---|---|---|---|---|
| Kotsovos et al. (Kotsovos) | S1 | 205 | 255 | 6.2 | 24.25 | 655 | 0.085 |
| Kotsovos et al. (Kotsovos) | S3 | 205 | 255 | 6.2 | 24.25 | 665 | 0.345 |
| Caldentey et al. (Caldentey, Lavaselli et al. 2013) | C1 | 255 | 455 | 5.6 | 33.95 | 555 | 1.075 |

## 8 Conclusions

This study investigated the potential of machine learning models to predict the punching shear strength of flat slabs. Three machine learning models have been built: Artificial Neural Networks (ANN), Particle Swarm Optimized Feed Forward Neural Network (PSO-FNN) and Bat algorithm optimized Feed Forward Neural Network (Bat-FNN). The performance of these models have been determined using Mean Square Error objective (MSE) function. In order to enhance the reliability of this study, design codes of American Concrete Institute (ACI), Eurocode and an analytical approach of compressive force path method has been used to determine the punching shear strength of flat slabs. These predictions are compared with the actual experimental values and results of empirical models. A database of square flat slabs consisting of 145 samples has been built from literature with parameters related to the slab depth, column dimension, shear span ratio of slab, yield strength of longitudinal steel, longitudinal reinforcement ratio, ultimate load carrying capacity and compressive strength of concrete.

The machine learning models performed training and testing of the dataset with greater accuracy, some of the hybrid models of neural networks can outperform standalone neural networks, other hybrid models and the empirical models of punching shear strength of flat slabs. Hybrid model of PSOFNN outperformed all other models including both the machine learning models and empirical models. This shows greater accuracy and optimization ability of PSOFNN for the punching shear strength of flat slabs and provides good agreement with the experimentally measured values of punching shear strength of flat slabs. The PSOFNN consistently decreased the Mean Squared Error (MSE) function, over iterations effectively optimizing the weights of the FNN. While BATFNN showed limited improvement in the objective function after the initial iterations indicating a plateau, in its optimization performance. Out of the seven datasets, the optimized PSOFNN model for the Square Concrete Slabs (SCS=1) exhibited the best performance. It achieved impressive metrics, with correlation coefficient, Mean Squared Error, and Mean Absolute Error values of 99.37%, 0.0275%, and 1.214%, respectively. It's important to note that the results obtained through PSOFNN were validated using the Nonlinear Finite Element Analysis (NLFEA) tool in ABAQUS. Consequently, PSOFNN offers a cost-effective and time-efficient alternative to experimental testing for estimating the behavior of flat slab-column connections in structural engineering applications.

The dataset utilized in this study is exclusively composed of square concrete slabs, thus confining the data collection to a particular slab type, which consequently accounts for the relatively modest dataset size, encompassing a total of only 145 samples. Although this dataset was sufficient for the machine learning models but future research could benefit from the larger datasets.

**Author Contributions**

All authors have read and agreed to the published version of the manuscript.

**Funding:** The authors received no external funding.

**Conflicts of Interest:** The authors declare no conflict of interest.